\documentclass[letterpaper, 10 pt, conference]{ieeeconf}  % Comment this line out if you need a4paper

\IEEEoverridecommandlockouts                              % This command is only needed if 
\usepackage{graphics} % for pdf, bitmapped graphics files
\usepackage{epsfig} % for postscript graphics files
\usepackage{mathptmx} % assumes new font selection scheme installed
\usepackage{times} % assumes new font selection scheme installed
\usepackage{amsmath} % assumes amsmath package installed
\usepackage{amssymb}  % assumes amsmath package installed
\usepackage{booktabs}
\usepackage[colorlinks = true]{hyperref}
\usepackage{float}
\usepackage{changepage}

\usepackage{xcolor}
\let\labelindent\relax
\usepackage{enumitem}

\usepackage{xspace}
\newcommand\OURS{SEaT\xspace}

\usepackage{cite}

\DeclareTextSymbolDefault{\DH}{T1}
\usepackage{multirow}

\title{\LARGE \bf TeleSnap: Scene-Centric Teleoperation with Action Snapping \\for 6DoF Kit Assembly}

\title{\LARGE \bf Scene Editing as Teleoperation: A Case Study in 6DoF Kit Assembly \vspace{-3mm}}

\author{ Yulong Li$^{* 1}$, Shubham Agrawal$^{* 1,2}$,  Jen-Shuo Liu$^{1}$, Steven K. Feiner$^{1}$, Shuran Song$^{1}$ \\

$^{1}$Columbia University $^{2}$Samsung AI Center NY \\

\href{https://seat.cs.columbia.edu/}{https://seat.cs.columbia.edu/}
 
\thanks{$^*$ indicates equal contributions}
}

\begin{document}

\maketitle
\thispagestyle{empty}
\pagestyle{empty}

\begin{abstract}
Studies in robot teleoperation have been centered around action specifications---from continuous joint control to discrete end-effector pose control. However, these ``robot-centric'' interfaces often require skilled operators with extensive robotics expertise.   
To make teleoperation accessible to non-expert users, we propose the framework ``Scene Editing as Teleoperation'' (SEaT), where the key idea is to transform the traditional ``robot-centric'' interface into a ``scene-centric'' interface---instead of controlling the robot, users focus on specifying the task's goal by manipulating digital twins of the real-world objects. As a result, a user can perform teleoperation without any expert knowledge of the robot hardware. 
To achieve this goal, we utilize a category-agnostic scene-completion algorithm that translates the real-world workspace (with unknown objects) into a manipulable virtual scene representation and an action-snapping algorithm that refines the user input before generating the robot's action plan.  
To train the algorithms, we procedurely generated a large-scale, diverse kit-assembly dataset that contains object-kit pairs that mimic real-world object-kitting tasks. Our experiments in simulation and on a real-world system demonstrate that our framework improves both the efficiency and success rate for 6DoF kit-assembly tasks. A user study demonstrates that SEaT framework participants achieve a higher task success rate  and report a lower subjective workload compared to an alternative robot-centric interface.%Experiment videos can be found at \href{https://seat.cs.columbia.edu/}{https://seat.cs.columbia.edu/}. %Code and data will be available. 
\end{abstract}

\section{Introduction}
\label{sec:intro}
% - internet delay
% - hard to control robot end effector
%Robot teleoperation becomes increasingly crucial to enable remote physical works (e.g., assembly, kitting, and packaging). 
The vast majority of robot-teleoperation research has focused on how to better specify robot actions: from  continuous joint control to discrete end-effector pose control. However, most of these “robot-centric” interfaces require skilled operators (with robotics expertise), complex input devices, or low-latency connections, which are hard to guarantee in practice. 
%In particular, the teleoperators who has the expertise and experience in both original tasks (e.g., assembly) and robot controls are extremely hard to find. 

% key ideas  
To address these issues, we propose the framework of \textbf{``Scene Editing as Teleoperation'' (\OURS)}, where the key idea is to transform the traditional \textit{robot-centric} interface into a \textit{scene-centric} interface---instead of specifying robot actions, users focus on specifying task goals by manipulating digital twins of real-world objects. As a result, non-expert users, users who have a high-level understanding of the task but no experience of working with the robot, can perform teleoperation without knowledge of the robot hardware, control mechanisms, or current state---users do not even see the robot during teleoperation.  
In addition, by removing the need of continuous control, the system is able to gracefully handle variable network latency. 

While \OURS is applicable for general ``object rearrangement'' tasks, we use 6DoF unknown object kit assembly as the case study in this paper. This task is selected because of its high requirements in precision and flexibility. Through this task, we hope to demonstrate the useful capabilities of \OURS that could not be achieved by either a traditional teleoperation system (struggles to produce precise actions in 6DoF space \cite{kent2020leveraging}) or an automated system (struggles to generalize to new objects and tasks \cite{devgon2021automating}).

While there are many existing ``scene editing'' tools for manipulating virtual objects  \cite{reinhart2009experimental,unity,solidworks}, the decisive challenge for our task is how to reliably translate between the real and virtual scene representations, specifically: 
\begin{itemize}[leftmargin=*]
    \item How to translate the realworld workspace filled with \textit{\textbf{unknown}} objects into an editable virtual scene.
    \item How to translate \textit{\textbf{imprecise}} user edits (i.e., objects' rearrangements) to the realworld with the robot's actions. 
\end{itemize}

\begin{figure}[t]
    \centering
    \includegraphics[width= 0.98\linewidth]{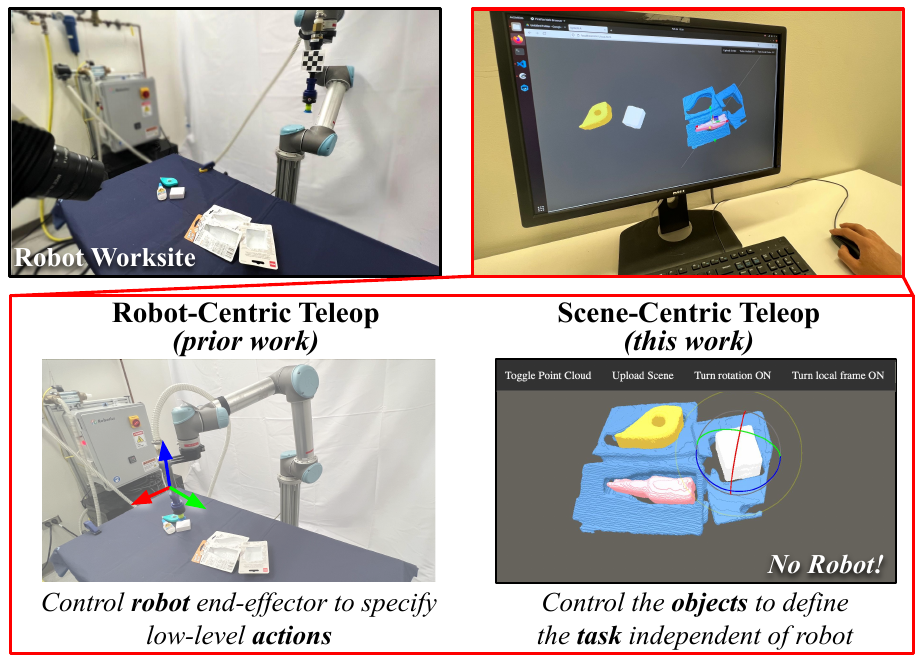} \vspace{-3mm}
    \caption{\textbf{Scene Editing as Teleoperation.} With a scene-centric interface, our framework allows the user to efficiently specify the task goal without expert knowledge of the robot hardware or control, making this framework accessible to non-expert users. By removing the need for continuous control, the system is able to gracefully handle variable network latency. } \vspace{-3mm}
    % new: https://docs.google.com/drawings/d/1pjSBoyH6hTmWPwLNqSi0jfYOGs3ZWoDKIsqUSm6GQT8/edit
    % old: https://docs.google.com/drawings/d/1PicvCidGSQMa5KuNgT5RBqjcuNxleQjfobKPtrkQbZA/edit
    \label{fig:overview}
\end{figure}

To obtain the digital twins of unknown objects,  we propose a category-agnostic scene-completion algorithm that segment and complete individual objects from depth images.
To handle imprecise user inputs, we propose a 6DoF action-snapping algorithm that automatically refines user inputs and corrects object-kit alignment using a 3D shape matching network.
Finally, virtual operations on object poses are translated by a sequence of robot actions generated by the robot planner. 
Learning from a large-scale kit-assembly dataset, our framework (both scene-completion and action-snapping algorithms) can generalize to unseen object-kit pairs, allowing quick adaptation to new assembly tasks.

In summary, our primary contribution is the framework of \OURS that allows non-expert end users to perform complex and precise 6DoF kit-assembly tasks over a high-latency internet connection. This framework is enabled by the following technical contributions: 
\begin{itemize}[leftmargin=*]
    \item A category-agnostic scene-completion algorithm that translates the real-world robot workspace (with unknown objects) into a virtual editable scene representation. 

    \item An action-snapping algorithm that automatically refines user inputs and improves object-kit alignment using a 3D shape matching network. 
    
    \item A large-scale kit-assembly dataset, KIT1000, that contains a diverse set of procedurally generated object-kit pairs that mimic real-world kitting tasks. This diverse training data allows the algorithm to generalize to new objects. 

\end{itemize}
Extensive experiments suggest that \OURS improves both the efficiency and success rate of 6DoF kit-assembly tasks, while achieving a lower subjective workload compared to an alternative robot-centric interface.
Please see our \href{https://seat.cs.columbia.edu/}{project website} for more system videos. Code and data will be made publicly available.

\section{Related Work}
\label{sec:related_work}

\textbf{Teleoperation.}
Early investigations in robot teleoperation focused on specifying a continuous motion trajectory \cite{direct08, directForce, Hayati1989, Oda1999, Michelman1994, leeper2012,kent2020,ciocarlie2012,gualtieri2016}, which often requires a low-latency connection between the teleoperator and robot or a complex input device for the operator. To reduce these requirements, other systems allow the operator to specify only the robot end-effector target poses \cite{gadre2019end, gossow2011interactive, kent2020leveraging,teach14}, and allow asynchronous execution to mitigate high communication latency. 
However, regardless of the levels of control, all these systems still focus on specifying the \textbf{robot's action}, requiring expert teleoperators with knowledge and intuition of the robot embodiment. For example, the user needs to understand the robot kinematics to specify a reachable and collision-free arm trajectory or understand the robot gripper mechanism to specify a valid grasp pose. Training human operators with this expertise can be expensive and difficult to scale. 
In contrast, our system focus on specifying the \textbf{task goal} regardless of robot hardware. This idea of task-driven teleoperation has been studied in simple scenarios such as point-goal navigation \cite{Lee14AAAI} or manipulation with known objects \cite{ciocarlie2012mobile}. However, how to enable precise and efficient task specification for a complex assembly task with unknown object parts is still an open research question, hence the focus of this paper.  

%\yulong{shuran:  add  more related work in this section, need review to see if it's sufficient}

%To ease the low-latency requirement, direct teaching systems require users to move the robot's links \cite{direct08, directForce}, manually perform the actions with mounted sensors \cite{direct08, directGlove}, or record human demonstrations for robot imitation \cite{directAsembly, directAsembly2}. The data collection process for these methods is physically demanding and time-consuming.
%Alternatives to direct teaching systems are teaching pendant systems that offer users a control system, typically with a GUI and on-screen buttons, to control the robot \cite{teach12, teach14, teach15}, which requires the user's proficiency with the control system. There are also many efforts to make the control system more accessible: for example, Lin \etal \cite{teach14} designs a teach pen as the motion capture system.

\textbf{Vision-based kit assembly.}
Traditional vision-based assembly approaches require strong prior knowledge of target objects (e.g., detailed CAD models) to perform object-pose estimation and motion planning \cite{Litvak18,devgon2021automating}. As a result, these approaches often cannot generalize to new objects without extensive data collection.  
Recent methods explore the idea of shape-informed assembly \cite{zakka2020form2fit, zeng2020transporter, devgon2021automating}, where the task of assembly is formulated as a shape-matching problem between the object and its target location. This formulation allows the algorithms to generalize toward unseen objects by directly analyzing their 3D geometry.
However, these algorithms are still limited to simpler tasks, such as 3DoF assembly \cite{zakka2020form2fit}, only predicting single object assembly \cite{devgon2021automating, zeng2020transporter}, only rotation prediction \cite{devgon2021automating} or require precise demonstrations on the exact object-kit pair \cite{zeng2020transporter}. While top-down kits (3DoF assembly) may seem ubiquitous, most do not have a flat bottom and hence cannot stand vertically on their own on an assembly belt. Handling multiple objects simultaneously is required for kitting tasks involving packaging multiple related objects together (e.g., toothpaste and toothbrush or bundle of pens).
% However, multi-object 6DoF kitting is not only challenging due to a large search space, but also has natural ambiguities that cannot be resolved by shape analysis alone (e.g., similar shape but the different texture or multiple matching candidates due to the symmetry).
Our approach is able to handle multi-unknown-object 6DoF kitting from imprecise user input, where user input helps reduce potential ambiguities and reduce search space, and the 3D shape-matching network further refines imprecise user input. 
%Moreover, our shape matching algorithm computes the correspondence based on the objects' completed 3D geometry instead partial depth observations \cite{zakka2020form2fit,zeng2020transporter}, which allows precise 6DoF pose refinement. 

%\todo{address here why kit-net's kitting algorithm is not enough or different: (1) they assume that posing the object at particular pose and opening the gripper will result in sufficient kitting. However, for most kits, you need to approach from particular direction. Different in terms of we orient 2d extruded top down kits. Also they train on real world dataset with 200 depth image pairs. Their translation prediction is heuristic based which highly limits the applicability} 

% Prior works in 6DoF kitting have handled these challenges by making several limiting assumptions such as known relative translation (e.g., KitNet \cite{devgon2021automating} predicts only rotation) or known object-kit pair during testing (e.g., the L-shape kit used in TransporterNet 6DoF kitting task ). Both methods fail to generalize to our scenarios (Tab. \ref{table:snapnet}).

\begin{figure}
\centering
\includegraphics[width=\linewidth]{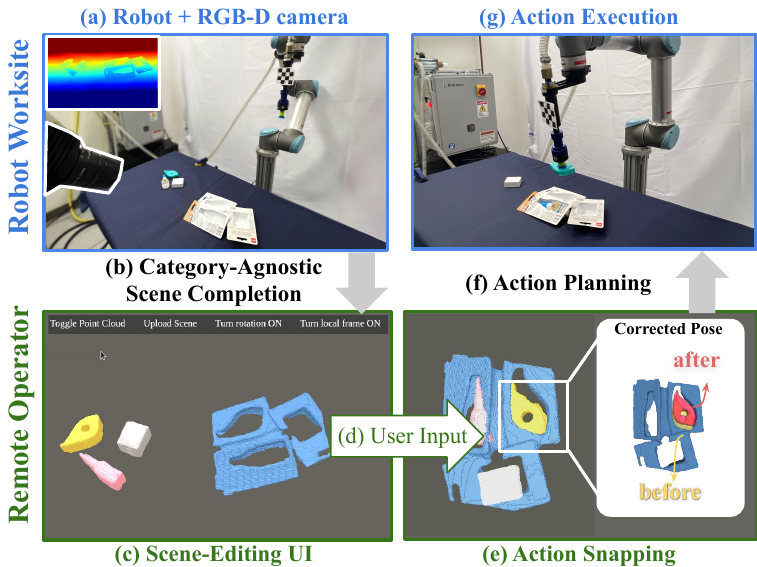} \vspace{-5mm}
% new: https://docs.google.com/drawings/d/18SvN0z1CBSXMKEK16uOaxS9OysDAE4dlq0KdxyY5W4w/edit
% old: https://docs.google.com/drawings/d/1IhQRrK_7BS9jpSXl06O2VA4VxwFRpDPdhK7p3TsbgKk/edit

\caption{\textbf{Overview.}  Given a depth image, the scene-completion algorithm converts the workspace into a virtual scene (a--b \S \ref{sec:completion}). The user then specifies a target object pose by editing the virtual scene using the 3D UI (c--d, \S \ref{sec:editing}).  Our action-snapping algorithm refines the object pose to improve object-kit alignment (e, \S \ref{sec:snapping}).  Finally, the system computes and executes the actions to assemble the objects (f--g, \S \ref{sec:planning}).}
\label{fig: system_pipeline} \vspace{-7mm}
\end{figure}

%training these algorithms still requires large of realworld data that might be difficult 

\textbf{Creating digital twins of 3D scenes.}
Many 3D scene-understanding algorithms have been developed to produce high-quality digital models of real-world environments for teleoperation.
These include algorithms for 3D object detection \cite{song2016deep,schwarz2018rgb,taylor2012vitruvian,guler2018densepose,brachmann2014learning,gupta2015aligning,guo2014scene,papon2015semantic,braun2016pose} and shape completion \cite{song2017semantic,dai2018scancomplete,xu2020learning,liang2021sscnav, liu2015deep,laina2016deeper}. 
Unlike traditional 3D scene-understanding tasks that focus on \textit{common} object categories (e.g., tables or chairs), in assembly tasks, a system often encounters a large number of new objects and parts that cannot be categorized into predefined categories. To address this issue, we propose a category-agnostic scene-completion algorithm that generalizes to unseen objects or parts without their 3D CAD model, allowing quick adaptation to new assembly tasks.
\begin{figure*}[t]
\centering \includegraphics[width=0.98\linewidth]{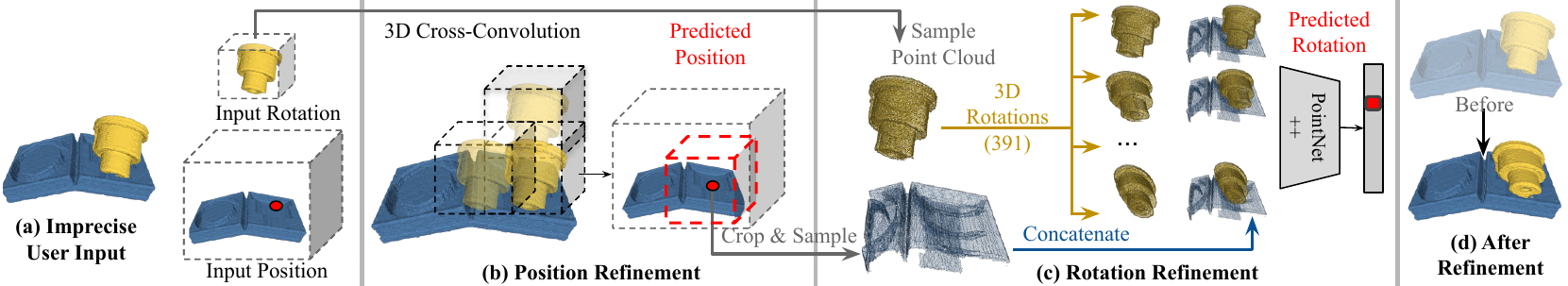} \vspace{-3mm}
\caption{\textbf{6DoF Action Snapping with SnapNet.} SnapNet uses 3D shape matching to refine the alignments between objects and their kits.  Given the user’s imprecise input (a), the algorithm first refines the object position by using a 3D cross-convolution network between the geometric features computed from the object and kit volume (b). The cross-convolution is computed only in the local area around the user inputs (b). The algorithm then samples point-clouds from the object volume and the cropped kit volume centered at the predicted position and predicts the refined rotation from 391 rotations using a PointNet++ based classifier (c). Finally the algorithm outputs the refined position and rotation as the target pose. }
\vspace{-7mm}
\label{fig: snapnet}
\end{figure*}

\section{Method: Scene  Editing  as Teleoperation}
\label{sec:approach}

We study the task of \textbf{6DoF kit-assembly with multiple unknown objects}.
To perform the task, the robot need to precisely place the object into their corresponding kit location with correct 6DoF poses. 
This task presents a set of unique challenges compared to general object rearrangement tasks:  
1) High precision requirement -- making it particularly challenging for human teleoperators with single view observation, hence, motivates our action snapping network with shape completed objects.  
2) Ambiguities in object-kit correspondence.  The ambiguities can be caused by similar or symmetrical shapes, requiring human inputs to disambiguate. 
3) Large search space --- compared to top-down kit-assembly tasks \cite{zakka2020form2fit}, the possible object poses in 6DoF is significantly higher, making uniform search approach impractical.
4) Despite the  ubiquity of the kit-assembly applications, a large-scale dataset is not yet available for this task, which is a key bottleneck for enabling learning-based approaches. 
In the following sections, we will discuss our approach to address above challenges.

\subsection{Category-Agnostic Scene Completion} 
\label{sec:completion}
Given a single depth image $I$ of the workspace with objects on one side and the kit on the other, the algorithm generates shape-completed geometries for individual objects using the following two steps: 

\underline{Object-Instance Segmentation:} 
The algorithm first detects and segments all object instances using SD-MaskRCNN \cite{danielczuk2019segmenting}: a variant of MaskRCNN \cite{he2017mask} using only depth for better sim2real generalization. Since the assembly task involves a large number of object parts that cannot be categorized into predefined categories, we train this algorithm in a category-agnostic manner with only a binary objectness label. 

\underline{3D Shape Completion:} 
Given an object's instance mask $M$ and the depth image $I$, the algorithm estimates the object's full 3D geometry. This shape-completion step provides two benefits: 1) it aids the user during teleoperation by better visualization of the objects and provides more context for successful kitting, and 2) it helps in achieving better action-snapping results as shown in Tab. \ref{table:snapnet}.

To perform 3D shape completion, we first transform partial object geometry information from masked depth image $MD = I \times M$ into a $128^3$ TSDF volume \cite{newcombe2011kinectfusion} representation $V_{partial}$  with voxel size $0.89$ mm. %We use the $MD$, known camera intrinsics, and camera extrinsics from calibration, to generate the input TSDF volume $V_{partial}$ of dimensions $(128 \times 128 \times 128)$ and voxel size $0.89$ mm around the center of the segmented point cloud. The voxel size was chosen to be smaller than the object-kit alignment error margin of $2.5$ mm. The volume dimension of $128 \times 0.89 = 11.4$ cm was chosen to be bigger than the object dimensions in our dataset. 
This volume is then fed into our shape-completion network $SC_{\theta}$ to obtain the shape-completed 3D volume $V_{completed}$. $SC_{\theta}$ follows a 3D encoder--decoder style architecture with skip connections \cite{xu2020learning}. The network is trained to minimize voxel-wise MSE loss. We train a separate network for kits with same architecture as for object shape completion. %For kit shape completion, we compute an input TSDF volume of dimensions $(400 \times 400 \times 256)$ and voxel size $0.89$ mm around the center of the kit workspace. We train a separate network for kits with same architecture as for object shape completion. \shubham{Please verify changes @shuran @yulong: too complex? unclear?}
%For the kit, since we assume only one (multi-)kit presents in the kit workspace, we don't need segmentation. We directly utilize the partial kit workspace volume as input to a separately trained shape completion network that predicts the completed kit workspace volume.

Both models are trained on the simulation data generated from objects and kits from our dataset (see \S \ref{sec: approach_dataset}) and then directly tested on unseen real world data.

\subsection{Scene-Editing  Interface \label{sec:editing}}
Given the 3D models for each object, the next step is to specify the task goal by changing their 3D poses in a virtual scene. This interface (Fig. \ref{fig: system_pipeline} c) is implemented as a 3D UI in a standard web browser using the three.js library \cite{danchilla2012three}. 
The user can observe the 3D scene from an arbitrary viewpoint and select, translate, and rotate individual objects. The user sends the target poses to the robot by clicking the \textit{Upload Scene} button. 
Our user study demonstrates that being able to directly manipulate objects at their target kits significantly reduces subjective workload as compared to traditional methods. Moreover, our interface does not require specialized hardware or a fast internet connection, making it accessible to common users (see video for interface demo). 

%and the user can manipulate individual objects to specify their target poses with following functionalities: (a) observing the completed 3D scene from arbitrary viewpoints (b) selecting, translating and rotating objects (implemented using \textit{TransformControls} class from three.js). To specify the target poses of objects, the user analyze the geometry of objects and kits, and place them one-by-one to the desired matching kit. User sends the target poses to the robot-site by clicking the \textit{Upload Scene} button. 

\vspace{-1mm}
\subsection{SnapNet: 6DoF Action-Snapping Network  \label{sec:snapping}}

Specifying perfect 6DoF kitting poses is challenging. As supported by our study, allowing users to be imprecise greatly reduces their mental burden and task time as they can roughly align an object near its respective kit.

To make use of imprecise user inputs, we designed 
the SnapNet algorithm (Fig. \ref{fig: snapnet} ) that refines the objects' pose based on their 3D geometry.  Concretely, the goal for SnapNet is to predict correct relative pose $T_{gt}$ between object and kit given input volumes of object  $V_{o}$, a kit $V_{k_{ws}}$, and user input $T_{user} \equiv (P_{user}, Q_{user}) \in SE(3)$. Here, we assume user input is within range: $max_{i \in \{x,y,z\}}|P_{i, user} - P_{i, gt}| < \delta_{position}$ and $Q_{user} . Q_{gt} ^{-1} < \delta_{orientation}$ where $T_{gt} \equiv (P_{gt}, Q_{gt})$ is the ground-truth kitting pose. 
We train our system to handle poses up to $\delta_{position} = 2.8$ cm error along each translational axis and $\delta_{orientation} = 27.5^\circ$ quaternion difference. %This tolerance is sufficient for our setup where simulation object-kit translational margin is $0.25$ mm and real-world object-kit translational margin is $1$ cm.

To reduce the combinatorial search space, SnapNet predicts translation and rotation sequentially, which reduces the search space from $O(\theta_{xyz} \times \theta_{rpy})$ to $O(\theta_{xyz} + \theta_{rpy})$ where $\theta_{xyz}$, $\theta_{rpy}$ represents discretization of translational and rotational search space.

\underline{Position prediction}:
Given $V_o$, $V_{k_{ws}}$ and $P_{user}$, the goal of position prediction is to infer $P_{snap}$. We first crop kit workspace volume $V_{k_{ws}}$ centered around $P_{user}$ and of size $(2\delta_{position})^3$ to receive $V_{k}$. We then encode $V_o$ and $V_{k}$ via object and kit encoders (fully convolutional neural networks) to obtain deep feature embeddings $\phi(V_o)$ and $\psi(V_k)$ respectively. The algorithm then computes cross-convolution between $\phi(V_o)$ and $\psi(V_k)$ by treating $\phi(V_o)$ as convolution kernel. The output shares the same size as kit features $\psi(V_k)$. $P_{snap}$ is chosen as position that corresponds to maximum feature correlation, i.e., $argmax$ of cross convolution output. Both encoders are trained jointly to minimize voxel-wise BinaryCrossEntropy loss with label 1 at $P_{gt}$ and 0 elsewhere.

% cross-convolve dense feature volumes of $V_{o}$ and $V_{kit_vicinity}$ obtained by two seperate deep convolutional networks. 
\underline{Rotation prediction}:
Given $V_k$, $V_o$, user orientation $Q_{user}$, and position prediction $P_{snap}$, the goal of the Rotation module is to predict $Q_{snap}$. Directly regressing quaternions \cite{devgon2021automating} fails to generalize (see Tab. \ref{table:snapnet}) and volume-based representations are susceptible to information loss under rotations.  %(c) sampling in SO(3) space is expensive
To address these issues, we
%choose to
use a \textbf{point-cloud--based} representation for rotation refinement. 
Using the refined position $P_{snap}$, $V_k$ is further cropped down at center with size $(128)^3$. Both $V_o$ and $V_k$  volumes are converted to point-cloud representation ($N_o = 2048$ and $N_k = 4096$ points $\in \mathbb{R}^3$ respectively) to support rotation operations. We uniformly sample $N - 1$ rotations within $\delta_{orientation}$ from the user input $Q_{user}$. $Q_{gt}$ is added to the set of rotations ($N=391$) during training. For each rotation $r$ in the set, we rotate the object point-cloud by $r$ and concatenate it with the kit point-cloud. An additional fourth dimension is utilized to distinguish between object $(1)$ and kit $(-1)$ points. A PointNet++ based encoder \cite{pointnet++} followed by fully connected layers is used to get binary classification score. We train the network using cross-entropy loss with $1$ for $Q_{gt}$ rotation and $0$ otherwise. 

All the modules are trained on the simulation data generated from objects and kits from our dataset (see \S \ref{sec: approach_dataset}) and then directly tested on unseen real world data.
\begin{figure}
    \includegraphics[width=0.98 \linewidth]{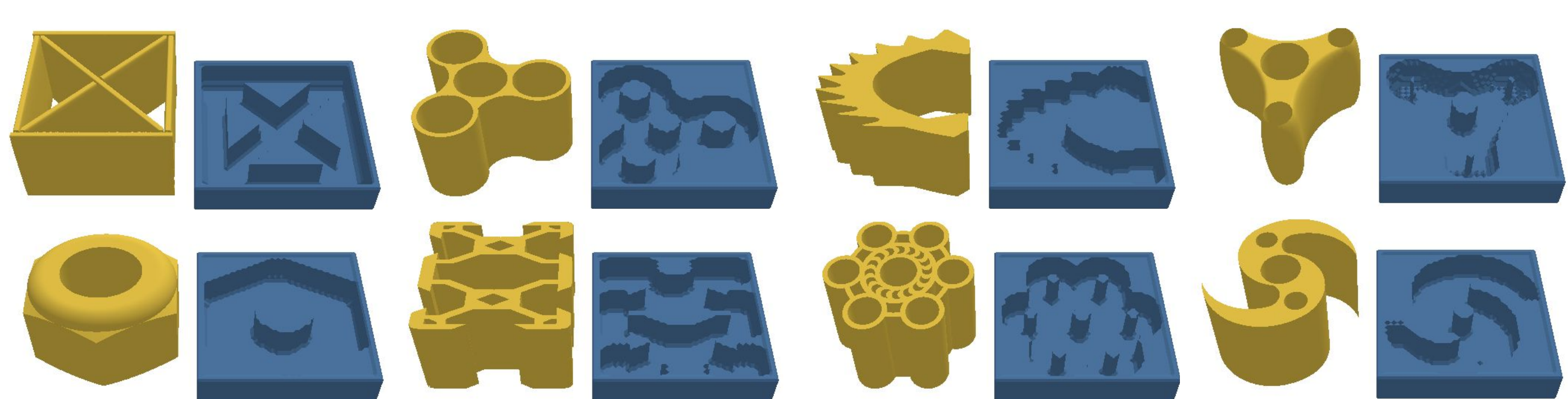}\vspace{-2mm}
    \caption{\textbf{KIT1000 Dataset.} Examples of objects and generated kits.}
    \label{fig:kit_sim}
    \vspace{-5mm}
    % https://docs.google.com/drawings/d/1m45GVutyP1mgSWTtZcFBdB2_al_3gidapwSSoYFG__0/edit
\end{figure}

\subsection{Robot Planning and Execution}
\label{sec:planning}
Picking and placing an object at specific goal pose is a challenging problem as the object may not initially be oriented such that the robot can grasp and then immediately place them in specific goal pose. Such manipulation systems are still an active research area \cite{wada2022reorientbot, chavan2018hand} and not the focus of this work. To test our system in real-world, we make a simplifying assumption that the object is top-down graspable, and the grasping surface is opposite to the kit insertion direction. No such assumptions are made for training and evaluation of scene completion and 6DoF pose prediction algorithms (Tab. \ref{table:snapnet}). To move the object from its current location to kitting location $^{robot}T_{snap}$, we pick the object via a suction-gripper--based top-down immobilizing grasp. The object is inserted into kit following a two-step primitive: (a) The robot first ``hovers" at some fixed height and final orientation above the kitting location defined as $^{robot}T_{hover} = ^{robot}T_{snap} \times ^{snap}T_{hover}$, where $^{snap}T_{hover} \equiv (^{snap}P_{hover} = [0, 0, 0.1]$ m, $^{snap}Q_{hover} = [0, 0, 0, 1])$. (b) The robot follows a straight-line path from $^{robot}T_{hover}$ to final pose $^{robot}T_{snap}$ before releasing the suction. More details on the grasp pose estimation and trajectory computation can be found on the \href{https://seat.cs.columbia.edu/}{webpage}.
%Note that the offset between object volume center and grasp position is ignored for writing simplicity in above calculation (see project webpage for details on grasp pose estimation and action planning). 

% Given an object to be moved at target position, we pick the object via a suction gripper based immobilizing grasp. Given the refined place pose $^{\text{world}}T_{place}$ from the SnapNet, we first orient the gripper at certain height above the place pose $ ^{place}T = [0, 0, 0.05]$, and then slowly move the gripper along the negative z axis in the place frame. More details on grasp detection in supplementary material.

\subsection{Dataset and Automatic Kit-Generation Procedure}
\label{sec: approach_dataset}
Despite the ubiquity of kits in the packaging and transport industry, most kits are manually designed and no large-scale object-kit dataset exists. Given a 3D object geometry, a typical kit (a) maximally confirms the object geometry and (b) allows the object to be inserted following a straight-line path at least along one direction. Our method neatly accounts for both of these: we capture an orthographic depth image of the object, which removes any artifacts that are not parallel to the insertion direction. The orthographic depth image is then converted to an occupancy grid. To allow some margin between kit and object geometry, the object 3D volume is then merged with replicas of itself after translating by margin distance along the horizontal direction. This creates a scaled version of the object geometry while preserving the centers of offset cavities. This scaled object geometry is then subtracted from the kit block to produce kit geometry. %(see Fig. \ref{fig:kit_procedure}) Kits generated using this procedure can be 3d-printed directly without any post-processing.  %\todo{should we move whole kit generation procedure to supplementary material?}

We use objects from ABC Dataset \cite{koch2019abc},  a large-scale CAD model dataset that contains a diverse set of mechanical parts. Each object is scaled to fit a $(5 cm)^3$ box and a corresponding kit is generated as described above (see Fig. \ref{fig:kit_sim}). To create 6DoF kits, we arbitrarily link 2--5 kits together using angle brackets with angles $\in [10^\circ, 45^\circ]$. We call this KIT1000 dataset and it will be made available.
\section{Experiments}
\label{sec:evaluation}
We first evaluate the action-snapping module (\S \ref{sec:eval_action_snapping}) followed by a full system evaluation on a real-world platform (\S \ref{sec:eval_real_world}) and a real-world user study (\S \ref{sec:eval_user_study}). 

\subsection{Action-Snapping Evaluation}
\label{sec:eval_action_snapping}
% We first evaluate the action snap algorithm and compare it with alternative approaches. We evaluate the algorithm performance under two conditions: (1) with user input and (2) without user input.
% When user input is available, we assume that the distance error is $ \leq 28 mm$, and rotational error is less than $\leq 27.5^{\circ}$. 

\textbf{Metrics:} We evaluate 6DoF pose prediction $T_{snap} \equiv (P_{snap}, Q_{snap})$ using two metrics: positional error $\delta_{\mathrm{pos}} =||P_{snap} - P_{gt}||_2$. Rotational error $\delta_{\mathrm{rot}}$ is computed as the geodesic distance $\arccos(2(Q_{snap} \cdot Q_{gt})^2 - 1)$. 

%\textbf{tasks:} We evaluate two tasks. The first task assumes the system receives a user input that orients the part in the vicinity of its correct pose. In particular, we assume that the distance between the ground truth position and the input position (the center of the kit volume) is at most $4o mm$, and that the rotational difference between ground truth orientation and the input orientation (the orientation of the input object volume) is at most $27.5^{\circ}$. Here we want to note that $40 mm$ is a very generous premise since it is at lest 40\% of the object size, and with sufficient visual cues, the client can usually does much better than that. This limits the search space of our model. For the second task, we assume no user inputs.

%Given a ground truth position $p$ and quaternion $q$ and corresponding predictions $\hat p$ and $\hat q$, the positional difference is the $L_2$ distance $\delta_p = ||p - \hat p||$ and the rotational difference is the geodesic distance $\delta_r = \arccos(2(q \cdot \hat q)^2 - 1)$. We report on the median differences for each model since median is robust to outliers compared with other statistics such as mean.

%\textbf{Baselines:} 

\begin{figure*}
    \centering
    \includegraphics[width=0.98\linewidth]{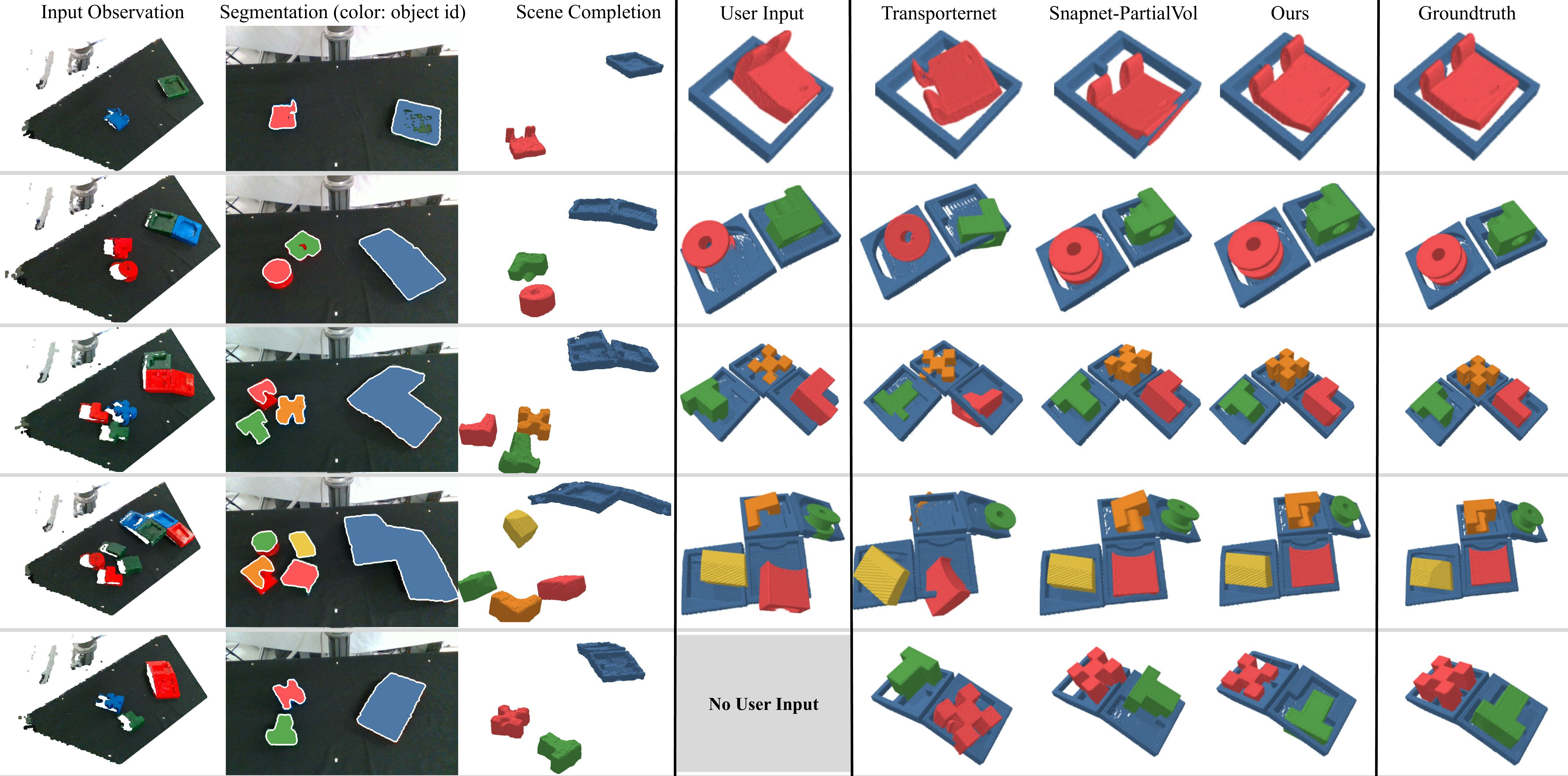}
    % https://docs.google.com/drawings/d/1xuE6Nhzjrh32Cgb2Me7IGti5fNYtvwaRSLa1vBHV5tM/edit
    \vspace{-3mm}
    \caption{\textbf{Comparisons to Alternative Approaches} We compare SEaT with 6DoF kitting baselines on novel object and kit geometries. TransporterNet fails to generalize to unseen object and kit geometries. SnapNet-PartialVol works for simple objects (row 2) but fails for objects with complex geometries (rows 3--4). When given no user input, both baselines frequently place objects at the wrong kits (row 5). In the last five columns, we use ground truth meshes to visualize poses.  For more results, see the \href{https://seat.cs.columbia.edu/}{project webpage.}} 
    \vspace{-3mm}
    \label{fig:result}
\end{figure*}

\textbf{Comparison with alternative approaches:}
We compare our algorithm with TransporterNet\cite{zeng2020transporter} and KitNet\cite{devgon2021automating}.
% - define with and without-user-input case.
Since both algorithms are trained without user input, we modify our algorithm to also work without user input: For position prediction, instead of cropping $V_{k_{ws}}$ around user input $P_{user}$, we directly use $V_{k_{ws}}$ as $V_k$. For rotation prediction, we uniformly sample $roll, pitch \in [-15^{\circ}, 15^{\circ}]$, and $yaw \in [-180^{\circ}, 180^{\circ}]$.
TransporterNet \cite{zeng2020transporter} consists of a pick and a place module. In our evaluation, we use the groundtruth pick position and retrain its place module with extensions to 6DoF actions. When user input is available, we filter out predictions that is far from provided pose, i.e., $T_{user} \pm (\delta_{position}, \delta_{orientation})$. KitNet \cite{devgon2021automating} predicts only the rotation of the object via regression, so there is no straightforward way to incorporate user inputs. Thus, we only evaluate the rotation predictions of KitNet without user input.  %For with-user-input case, since user provides both position and orientation it was ambiguous to use only position. Hence, we only evaluate their model for without-user-input rotation prediction.

% Since \textit{KitNet} \cite{devgon2021automating} does not handle position prediction,  we only compare for the rotational error with no user inputs.
% \textit{TransporterNet}  \cite{zeng2020transporter} infers 6DoF position and orientation from RGB images. When the user input is available, we will remove the predictions outside the plausible range.

Tab. \ref{table:snapnet} shows that both baselines fail to give accurate predictions. We hypothesize that without full geometry estimation, they do not have enough information to infer a 3D pose. By leveraging full 3D geometry and efficiently searching the SE(3) space, our model outperforms the baselines both with and without user input. \newline

\begin{table}[t]
\centering
\caption{Action-Snapping Results and Comparison} \vspace{-3mm}
\begin{tabular}{r|cc|cc}
\toprule
\multirow{2}{*}{} & \multicolumn{2}{c|}{With user input} & \multicolumn{2}{c}{Without user input} \\
& $\delta_{\mathrm{pos}}$(mm) & $\delta_{\mathrm{rot}}$(deg) &  $\delta_{\mathrm{pos}}$(mm) & $\delta_{\mathrm{rot}}$(deg)          \\ 
 \midrule
KitNet \cite{devgon2021automating}   & -        &     -    &   -   & 49.2         \\
TransporterNet  \cite{zeng2020transporter} & 15.3 & 18.3 & 41.5 & 45.1 \\ 
\midrule
SnapNet-PartialVol &  5.1   &   5.7   &   49.4   &   53.2  \\  
SnapNet (Ours) &  \textbf{3.9}   &   \textbf{4.9}   &   \textbf{10.8}   &   \textbf{29.6} \\ 
\midrule
SnapNet-GTVol  &  3.7   &   4.61  &   8.1    &   28.9   \\   
\bottomrule
\end{tabular}
\vspace{-3mm}
\label{table:snapnet}
\end{table}

\vspace{-3mm}\textbf{Effects of shape completion:}
To study the effect of shape completion on action snapping, we compare our approach without this step. \textit{SnapNet-PartialVol} uses partial volume $V_{partial}$ to perform shape matching.  
Tab. \ref{table:snapnet} shows that our model \textit{SnapNet} achieves better performance than  \textit{SnapNet-PartialVol}. We believe that this is because partial volumes lack of clear and precise object boundaries that shape matching crucially depends on. With ground-truth shape, \textit{SnapNet-GTVol} can further improve action-snapping performance. This result indicates that the scene-completion module is essential for achieving accurate action snapping. %We believe it successfully enables the model to more easily differentiate a completed shape between different kits, as seen in Fig. \ref{fig:result}, which the model fails with raw shapes.
 
 \begin{figure}[t]
    \centering
    \includegraphics[width=0.98\linewidth]{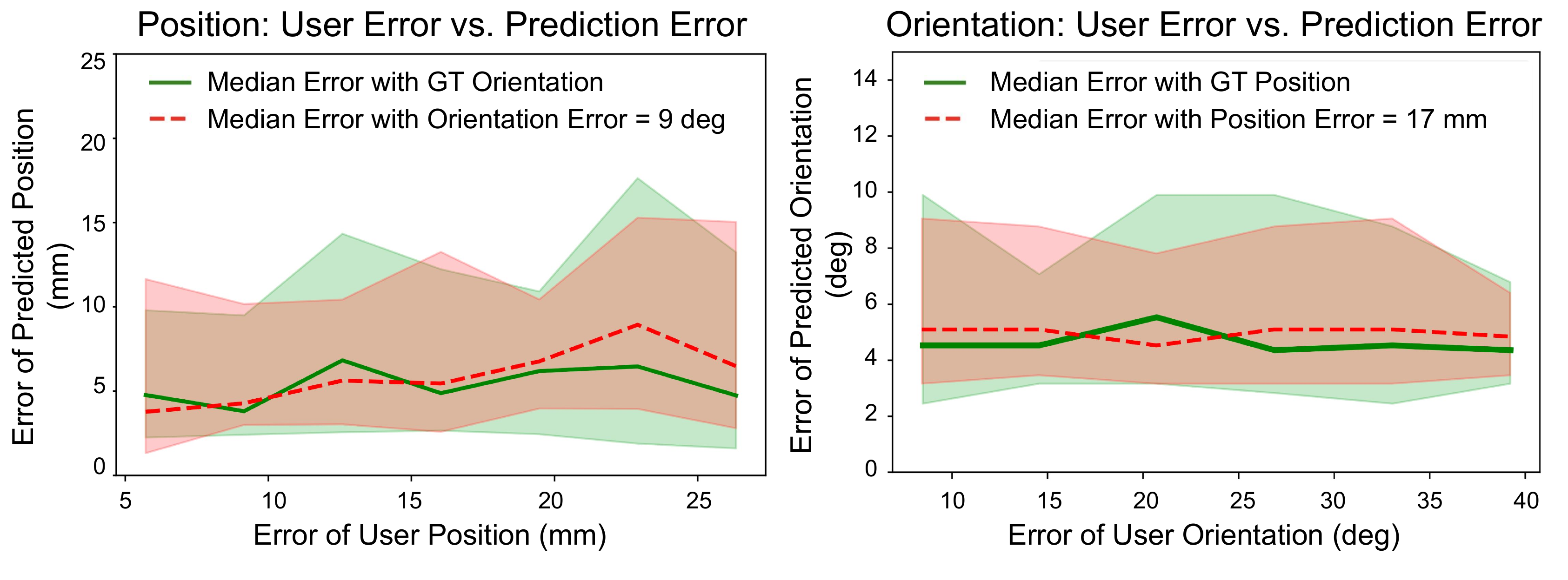} 
    \vspace{-2mm}
    \caption{\textbf{Robustness to User Input} with $[20, 80]$ percentile region shaded. The left graph shows an analysis of error in position prediction, keeping error in user orientation fixed. As user position error increases, SnapNet maintains its low prediction error. Moreover, even with a large error in user orientation (dotted-red), SnapNet can predict position with low error. Similar results for predicted orientation, keeping the error in user position fixed, are shown on the right.}
    \vspace{-7mm}
    \label{fig:robustness}
    % https://docs.google.com/drawings/d/1Umgh_XY-tgDEPHWPIVeYX1X0_oVRiFcLyAIU2tTqL1w/edit
\end{figure}

\textbf{Robustness against user errors:} We also test the algorithm's robustness to different levels of user-input error. For a controlled experiment, we analyze error in position and rotation prediction one-by-one by keeping the error in user orientation and user position fixed respectively.
Fig. \ref{fig:robustness} breaks down the performance of our model by plotting prediction errors $\delta_{\mathrm{pos}}, \delta_{\mathrm{rot}}$ against user-input errors. The plot shows that as user error increases, the model error remains roughly constant, demonstrating the robustness of the algorithm. %The sub-linear relationship shows that our model is robust against large user errors.
%

 %We observe that the main error the network makes is predicting a wrong piece of kit. %which often leads to wrong orientation prediction. %Our result encourages a promising direction for automatic vision-based assembly. 

\begin{table}[t]
\centering
\setlength\tabcolsep{3.5 pt}
\caption{System evaluation on the real-world dataset} 
\vspace{-3mm}
\begin{tabular}{c|cc|cc|cc}
\toprule
 \multicolumn{1}{c|}{Segmentation} & \multicolumn{2}{c|}{Obj. Completion} & \multicolumn{2}{c|}{Kit Completion} & \multicolumn{2}{c}{Action Snapping} \\
mIoU & mIoU & Chamfer & mIoU & Chamfer & pos & rot \\ 
 \midrule
%SegSc+TransporterNet  \cite{zeng2020transporter} & - & - & - & - &  - & 20.2 & 14.2 \\  \midrule
%SegSc+SnapNet-PartialVol & - & - & - & - &  - & 8.0 & 9.5 \\ 
\textbf{69.1}\%  & \textbf{92.4}\% & \textbf{6.3} mm & \textbf{99.1} \% &  \textbf{8.0} mm & \textbf{7.2} mm & \textbf{6.0}$^{\circ}$ \\ 
\bottomrule
\end{tabular}
\vspace{-5mm}
\label{table:real_eval}
\end{table}
% \vspace{-2mm}
\subsection{System Evaluation on Real-World Platform}
\label{sec:eval_real_world}
Finally, we evaluate our algorithm on a real-world platform using a UR5 robot, an XYZ Robotics suction gripper \cite{xyzRobotics}, and a calibrated Intel RealSense D415 RGB-D camera. To account for RealSense camera precision ($5$ mm depth error \cite{realsensePrecision}, for pick-place task, the error would be $10$ mm), we 3D-printed the kits from our test set with a larger object-kit margin of $1$ cm as compared to $2.5$ mm margin in simulation. 

For systematic evaluation, we collect and label $23$ scenes ($7$ of 1-kit, $7$ of 2-kit, $4$ of 3-kit, and $5$ of 4-kit tasks), with ground-truth object target poses. We directly tested all our models (trained on simulation) with this real-world benchmark. To eliminate small holes in shape completed object volumes $V_{completed}$ due to sensor noise in input $V_{partial}$, we extend all the object voxels till the ground plane. To mimic user input, we randomly sample position and orientation in the vicinity ($\delta_{position}, \delta_{orientation}$) of the ground-truth pose. 
Fig. \ref{fig:result} shows qualitative results on this real-world benchmark.  
Tab. \ref{table:real_eval} shows quantitative results for each individual component. The resulting average position and rotation error are comparable with the algorithm's performance in simulation (Tab. \ref{table:snapnet}). 
Moreover, our model has similar level performance on training and test dataset with unseen shapes, which shows that our model is generalizable by leveraging a large simulated dataset.

\begin{figure}
    \centering
    \includegraphics[width=0.98\linewidth]{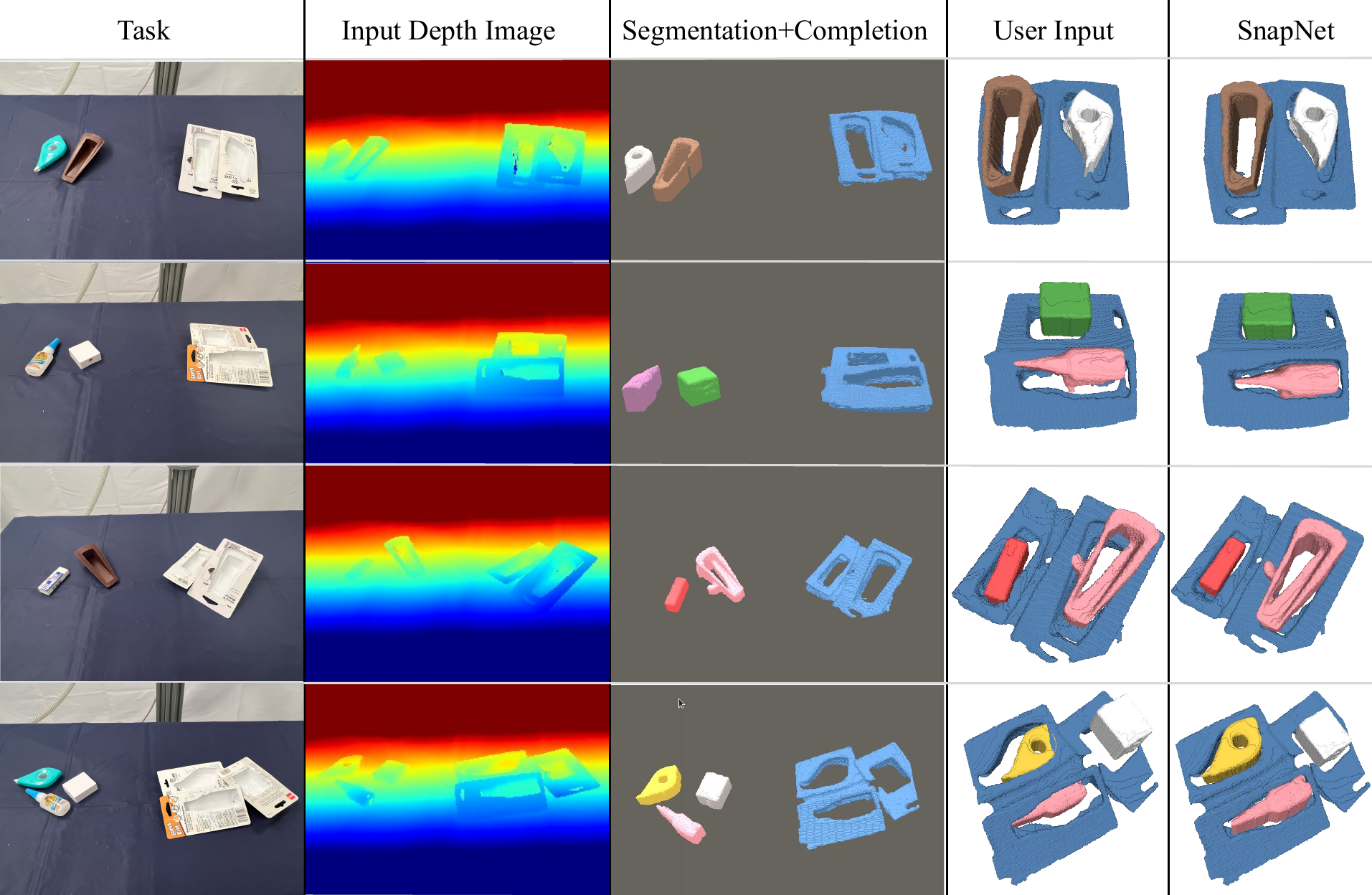} \vspace{-3mm}
    \caption{\textbf{Qualitative results on real-world kits.} See video for more results.} 
    % https://docs.google.com/drawings/d/1RQdI9LGcu9V5xWoAGZgHRHRL0sO9uncRD-U8QdtPdJA/edit
    \vspace{-5mm}
    \label{fig:real}
\end{figure}

In addition to 3D printed objects, we also evaluate the system on real-world object-kits (Fig. \ref{fig:real_kits}-bottom). Since these kits have a tighter object-kit margin, we use Photoneo Scanner with higher depth precision of $0.5$ mm \cite{photoneoPrecision}. Fig. \ref{fig:real} shows the qualitative evaluation. We refer readers to supplementary video for real-world demonstration of our system.
% \begin{figure}[t]
% \centering \includegraphics[width=0.98\linewidth]{figures/user-study.pdf} \vspace{-3mm}
% \caption{Quantitative and qualitative results from user-study. Error bars show standard error} \vspace{-3mm}
% \label{fig: user-study-results}
% % https://docs.google.com/spreadsheets/d/18nsCsFTydU_MMkXlawMpiRRuSzKWkYBmLic6qvp4sLc/edit?usp=sharing
% \end{figure}

\subsection{User Study on Real-World Platform}
\label{sec:eval_user_study}
Our user study aims to test the hypothesis that the \OURS interface would be easier to use than traditional teleoperation interfaces. We conducted a user study, approved by our institution's IRB with 10 non-expert users.    %We measured user task related times, success rate, system usability, and subjective cognitive workload for both interfaces.

\begin{figure}[t]
    \centering
    \includegraphics[width=0.97\linewidth]{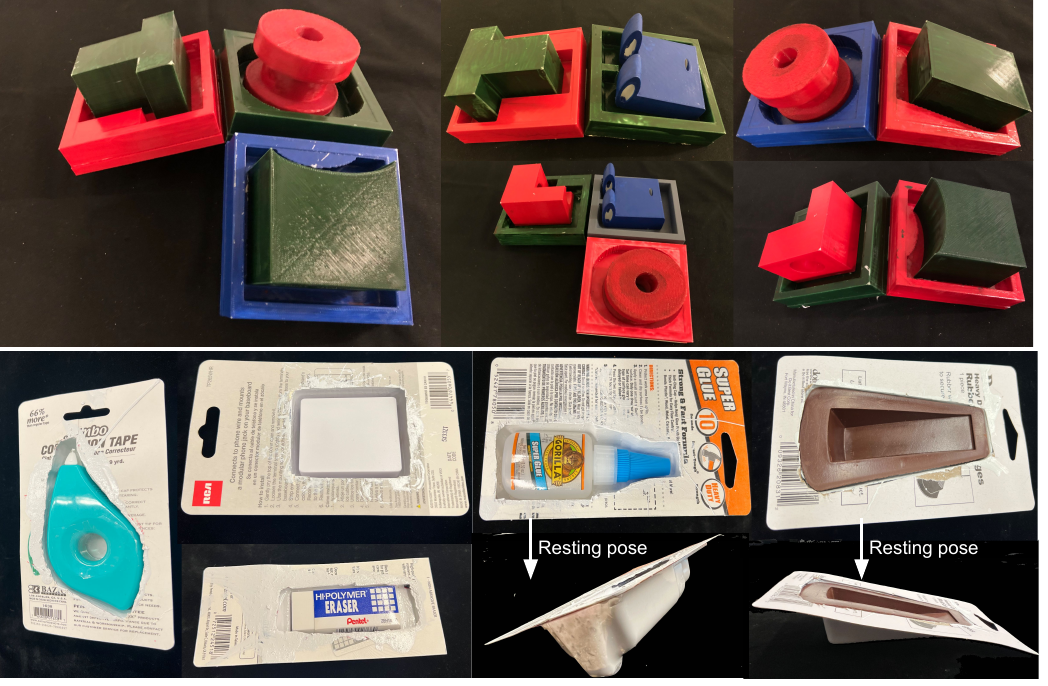} \vspace{-3mm}
    \caption{\textbf{Kits for real-world experiments.} Top: 3D-printed kits from test dataset are connected at arbitrary angles to create 6DoF kits. Bottom: real-world kits. Arrows show the resting pose for a few kits which require non top-down object insertion.}\vspace{-2mm}
    \label{fig:real_kits}
    % https://docs.google.com/drawings/d/1oS_GqmqnwhTiiYIK4Jr2lt95NUI-Z43E-h4lbfWxT6Q/edit?usp=sharing
    % with red-border: https://docs.google.com/drawings/d/1nQ010WBXVd31BBI8Khq6-wnCjoIZ7xgLd2PB781Ihwc/edit?usp=sharing
    \vspace{-3mm}
\end{figure}

\textbf{Task and Procedure:}
% , where an n-kit task means that there were n kits attached to each other at arbitrary angles to create one 6DoF kit.
Participants completed four kit-assembly tasks per interface (two 2-kit and two 3-kit tasks). For each $n$-kit task, we randomly attached $n$ kits from a set of six unseen 3D-printed kits using randomly chosen angle brackets $\{10^{\circ}, 20^{\circ}, 30^{\circ}\}$ (see Fig \ref{fig:real_kits}). %For each task, participants were asked to assemble all objects in the respective kits.
The study used a within-subjects design, where all participants performed both tasks using both interfaces in random order. Participants performed the 2-kit tasks first and then the 3-kit tasks for each interface.

\textbf{Comparisons:} 
We compared with EE-Control, a representative teleoperation interface where a user can specify 6DoF pick-and-place pose of the end-effector on the point-cloud representation of the scene. In the EE-Control interface, the user specifies a single pick-and-place pose followed by robot execution. Once the robot executes, the user scene is updated with the new scene and the user repeats the process. In  \OURS, the user specifies the goal poses of all objects at once.

% \textbf{Procedure:} 
% The study used a within-subjects design where all participants performed both tasks using both interfaces with random order. Experimenters counterbalanced the order of the two interfaces across study participants. For each interface, participants performed first the 2-kit task and then the 3-kit task. %More details can be find in \href{https://seat.cs.columbia.edu/}{project website}. 

%Participants began the study by reviewing the informed consent information. After they agreed to participate, the experimenters explained the two tasks. The users were randomly selected to first complete the tasks using either the SEaT or EE-Control interface. For each interface, users were first taught the controls. For each task, we recorded three metrics described below in \textit{Dependent Measures} section. After completing both the tasks, users filled out all usability and workload surveys for that interface.

\textbf{Dependent Measures:} Our objective dependent measures were  \textit{a. Success rate}: the number of kits successfully assembled over the total number of kits, \textit{b. specification time}: the time the user spent interacting with the interface for specifying goals, and \textit{c. execution time}: the total system time minus the specification time.  We also had a subjective dependent measure \textit{d. unweighted NASA Task Load Index (NASA-TLX)} \cite{hart2006nasa}, which includes values for MentalDemand, PhysicalDemand, TemporalDemand, Performance, Effort, and Frustration. 
Since a user is allowed to operate on the updated scene in the EE-Control interface, in theory they can always assemble all the objects given infinite time. Therefore, for both interfaces, a user can only start an update for an $n$-kit task if the time already spent is less than $n$ minutes. Users are informed about this time limit beforehand. We hypothesized that each of these dependent measures would differ between the SEaT and EE-Control interfaces.

%We hypothesized that SEaT would perform better than EE-control in terms of success rate, specification time, and mental demand.
% here will come all the details.

% \textit{f. Hypotheses}\\
% Overall, we expected users to perform better (lower task completion times, higher task success rate, lower work-load, higher usability and preference) on the SEaT interface than the EE-Control interface in both the tasks.\\
% \textbf{H1}: Users will quantitatively perform better on the SEaT interface than the EE-Control interface for completing both the tasks. Better performance is characterized as i) lower task specification time ii) lower total system time iii) higher task success rate.\\
% \textbf{H2}: Users will report higher subjective impressions of the SEaT interface than the EE-Control interface for completing both the tasks. Better subjective performance is characterized as i) lower reported workload and ii) higher usability.

% \begin{figure}
%     \centering
%     \includegraphics[width=0.98\linewidth]{figures/Telesnap_ user_study.pdf} \vspace{-3mm}
%     \caption{Comparison of Dependent measures ($\pm 1$ SE). For NASA-TLX, lower values suggest lower demand or better performance. }\vspace{-5mm}
%     \label{fig:user_study_results}
%     % https://docs.google.com/drawings/d/1ReGb9NOk0c8qf1pmOAftkR2zuszLBPhsK4r-oq59aAM/edit
%     \vspace{-0.5mm}
% \end{figure}

\textbf{Results:} 
We evaluated the hypotheses for significance with $\alpha = .05$. While the \textit{execution time} of \OURS is longer ($+12s$) due to model inference, the users spent significantly shorter \textit{specification time} ($-27s, p<.001$), and achieved significantly higher \textit{task success rate} ($+33.1\%, p<.001$). 
For subjective measures (\textit{NASA-TLX}), the participants reported significantly lower MentalDemand ($-39.2\%, p=.003$), lower TemporalDemand ( $ -43.1\%, p<.001$), lower Effort (  $-32.0\%,  p=.002$), and lower Frustration ($-40.7\%, p=.024$).  The reported differences in PhysicalDemand and Performance are not significant between these two methods. 
 
The shorter specification time and lower mental load of  \OURS indicates a potential possibility of using \OURS to simultaneously operate multiple robots. In this case, a user can continue specify tasks (for another robot) during model inference and robot execution time, which will further improve the system's overall efficiency. 
 
   %Please see \href{https://seat.cs.columbia.edu/}{project webpage} for separate \textit{NASA-TLX} scores and further analysis. 

 %\textit{(a) SEaT has a significantly higher successs rate with  \textit{(b}} the user spending significantly less time on averageg. SEaT \textit{(c)} has slightly longer execution time due to model inference, but  \textit{(d)} has lower load according to NASA load index \cite{hart2006nasa}.

% \begin{table*}[ht]
% \label{table:userstudy}
% \centering
% \caption{User Study Results} \vspace{-3mm}
% \begin{tabular}{c|c|cc|cccccc}
% \toprule
%  & Success Rate & Spec Time & Execution Time & \multicolumn{6}{c}{NASA Load Index} \\ 
% & (\%) & (sec) & (sec) & Mental & Physical & Temporal & Performance & Effort & Frustration \\
% \midrule
% SEaT       & 62.3 \pm 7.6  & 48.9 \pm 3.7 & 39 &  3.0 \pm 1.2 & 2.1\pm1.1 & 3\pm1.3 & 2.6\pm1.4 & 3.3\pm1.5 & 1.9\pm1.1 \\
% EE-Control & 29.2 \pm 20.0  & 77.8 \pm 9.7 & 27  & 4.9 \pm 1.2 & 2.6\pm1.4 & 5.4\pm1.2 & 3.8\pm1.3 & 4.9\pm1.2 & 3.1\pm1.9 \\
% \bottomrule
% \end{tabular}
% \end{table*}

% p<xx, Cohen's d=xx

%4) Order effects: Additionally analyses were conducted to test whether the order in which participants interacted with each interface biased the results. (xxx)

% Metrics:
% - Total time to complete the task
% - Accuracy of kitting

% \textbf{Setup:} 
% - how many task, how many user study  
% - describe tasks 

% \textbf{Metrics:} 
% time it takes to complete task, success rate 

% \textbf{results:} 
% - with and without action snapping 
% - without human input
% - controlling ee v.s. object: analysis failure cases 

\vspace{-2mm}
\section{Conclusion}
% A conclusion section is not required. Although a conclusion may review the main points of the paper, do not replicate the abstract as the conclusion. A conclusion might elaborate on the importance of the work or suggest applications and extensions.
We introduced ``Scene  Editing as Teleoperation'', which allows non-expert end users to perform precise multi-unknown-object 6DoF kitting tasks. Experiments demonstrated that SEaT improves efficiency, success rate, and subjective workload for 6DoF kit-assembly tasks.

% Other ideas:
% - doing without user input. 
% - Planning algorithm.
% - Extending this interface to other tasks.
Since our teleoperation interface assumes rigid objects, it cannot be directly applied to tasks involving articulated objects (e.g., opening a drawer). It would be interesting to discover articulation via RGB-D images \cite{gadre2021act, xu2021umpnet} and integrate it with our system. Planning the grasp and a set of sequential 6DoF robot actions for general 6DoF kitting tasks would also be an interesting future direction, where the robot might need to plan a place-driven grasp \cite{fang2020learning} or reorient the object before kitting \cite{chavan2015two}.  

% Many assembly tasks can be seen as generalization of 6DoF kit assembly task. For example, inserting headphone/usb in a computer, bolt in a nut, key in a lock, aligning screw driver with the hex-bolt etc. are primarily driven by shape matching. It would be interesting to explore their unique challenges and how well our algorithms generalizes to these different types of ''kits".

%\input{text/acknowledgement}
\newpage
\bibliographystyle{references/IEEEtran}
\bibliography{references/IEEEabrv,references/refs}

% \newpage
% \input{text/appendix/appendix.tex}
%\input{text/appendix}

% To view examples from ieee template, uncomment the following two lines
% \newpage
% \input{text/ieee_examples}

\addtolength{\textheight}{-12cm} 

\end{document}